\documentclass[10pt, conference]{IEEEtran}
\IEEEoverridecommandlockouts
\usepackage[T1]{fontenc}
\usepackage{cite}
\usepackage{amsmath,amssymb,amsfonts}
\usepackage{algorithmic}
\usepackage{graphicx}
\usepackage{epstopdf}
\usepackage{textcomp}
\usepackage{xcolor}
\usepackage{booktabs}
\usepackage{pgfplots}
\usepackage{pgfplotstable}
\usepackage{xurl}
\usepackage{multirow}
\usepackage[most]{tcolorbox}
\usepackage{enumitem}
\usepgfplotslibrary{groupplots}
\pgfplotsset{compat=newest}

\def\BibTeX{{\rm B\kern-.05em{\sc i\kern-.025em b}\kern-.08em
    T\kern-.1667em\lower.7ex\hbox{E}\kern-.125emX}}
\pgfplotsset{compat=1.18}
\begin{document}



\title{Cross-Domain Data Selection and Augmentation for Automatic Compliance Detection}

\author{

\IEEEauthorblockN{Fariz Ikhwantri\IEEEauthorrefmark{1}, Dusica Marijan\IEEEauthorrefmark{1}} \IEEEauthorblockA{\IEEEauthorrefmark{1}\textit{Validation Intelligence for Autonomous Software Systems} \\\textit{Simula Research Laboratory} \\ Oslo, Norway \\ fariz@simula.no; dusica@simula.no}

}

\maketitle

\begin{abstract}
Automating the detection of regulatory compliance remains a challenging task due to the complexity and variability of legal texts. Models trained on one regulation often fail to generalise to others. This limitation underscores the need for principled methods to improve cross-domain transfer. We study data selection as a strategy to mitigate negative transfer in compliance detection framed as a natural language inference (NLI) task. Specifically, we evaluate four approaches for selecting augmentation data from a larger source domain: random sampling, Moore-Lewis’s cross-entropy difference, importance weighting, and embedding-based retrieval. We systematically vary the proportion of selected data to analyse its effect on cross-domain adaptation. Our findings demonstrate that targeted data selection substantially reduces negative transfer, offering a practical path toward scalable and reliable compliance automation across heterogeneous regulations.
\end{abstract}

\begin{IEEEkeywords}
Requirements engineering; Regulatory compliance; Transfer learning; Data augmentation; Dataset selection.
\end{IEEEkeywords}


\section{Introduction}





Ensuring that software data and privacy protection comply with regulatory requirements is a critical yet resource-intensive task~\cite{saeidi-etal-2021-cross, castellanos2022compliance, duvall2007continuous}. In the context of the General Data Protection Regulation (GDPR), organisations must validate that Data Processing Agreements (DPAs) align with legal provisions governing the collection, processing, and protection of personal data. Traditionally, this validation process requires extensive legal and technical expertise, making it costly and time-consuming without automation.

To reduce this burden, researchers have explored automated approaches to compliance detection. Yet, developing models that can handle the complexity and variability of legal texts across domains remains a major challenge. Regulations differ not only in terminology but also in structure, intent, and reasoning style, making cross-domain generalisation particularly difficult.

\begin{figure*}[ht]
    \centering
    \includegraphics[width=0.7\linewidth]{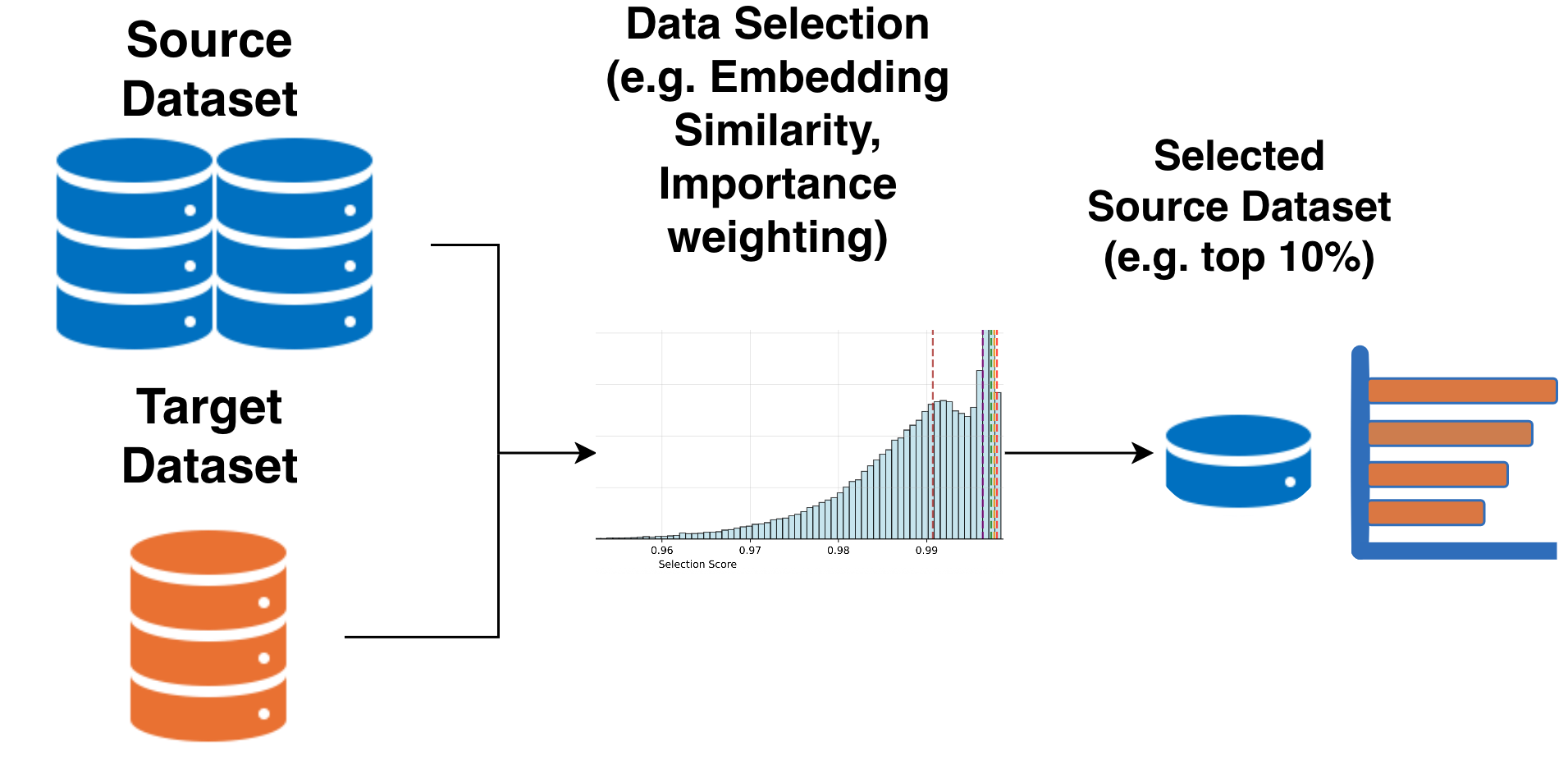}
    \caption{Given a large source domain (e.g. GDPR) and a smaller target domain (e.g. HIPAA), we evaluate several data selection methods to select a subset of the source domain data to prevent negative transfer.}
    \label{fig:data-selection-pipeline}
\end{figure*}

In single domain compliance detection, early work has relied on interpretable rule-based graph matching~\cite{10167495}, which provides transparency but struggles with linguistic variability, ambiguous legal phrasing, and limited scalability to new regulatory domains. More recent studies have reframed compliance as a text classification problem, applying statistical machine learning and language models to identify relevant clauses~\cite{azeem2024multi}. Another study proposed verbalising requirements as a class~\cite{Fazelnia2024LessonsFT} and compliance verification~\cite{10756275} as natural language inference (NLI) tasks, which offers a more structured formulation and some degree of data-driven explainability. Recent studies~\cite{ikhwantri2025explainable, etezadi2025classificationpromptingcasestudy} have leveraged Natural Language Inference (NLI) by framing requirements as premises and contractual clauses or policy statements as hypotheses, enabling systematic reasoning over regulatory alignment. 

While effective in single-domain settings, these approaches face a persistent bottleneck: the scarcity and domain-specificity of annotated datasets. Resources such as the GDPR dataset provide valuable supervision, but their limited size restricts model performance and their narrow scope hampers generalisation beyond a single regulatory framework.

In a cross-domain setting, J. Sun et al.,~\cite{sun-etal-2025-compliance} propose a retrieval-augmented generation framework to improve clause matching and explanation quality. Another recent study by Etezadi~\cite{etezadi2025classificationpromptingcasestudy} examines classification and prompting strategies for the traceability of legal requirements. They evaluate models on multiple domains, such as the General Data Protection Regulation (GDPR) and the Health Insurance Portability and Accountability Act (HIPAA), in the health domain. While these works highlight promising directions, they primarily focus on multiple-domain evaluation rather than cross-domain transfer, where models trained in one regulatory domain are applied to another regulatory domain.

A natural solution is to augment scarce datasets with additional training data, either by generating synthetic examples or transferring data from related domains. However, prior studies on transfer learning warn that not all data is equally useful~\cite{Wang2018CharacterizingAA}. Introducing poorly aligned samples can cause negative transfer, where performance degrades rather than improves when moving across domains~\cite{meftah-etal-2021-hidden}. This challenge highlights the need for principled data selection strategies that ensure augmentation enhances both predictive performance and explanation quality in compliance detection.




In this work, we systematically investigate cross-domain data selection for compliance detection. Figure~\ref{fig:data-selection-pipeline} illustrates the abstract data selection methods that were used in our study. In this work, we make the following contributions:

\begin{itemize}
    \item \textbf{Systematic study of cross-domain data selection.} We investigate the impact of augmenting compliance detection with source-domain data when transferring from GDPR to HIPAA.  

    \item \textbf{Evaluation of four selection strategies}. We compare two baseline random sampling and full augmentation with three methods: (1) Moore--Lewis cross-entropy filtering~\cite{moore-lewis-2010-intelligent}, (2) importance weighting, and embedding-based retrieval (DSIR)~\cite{xie2023data} as candidate approaches for source data selection.  

    \item \textbf{Analysis of scaling effects}. We examine how performance changes under different selection ratios, highlighting when and why negative transfer emerges and how targeted selection mitigates it.  

    \item \textbf{Empirical insights for compliance automation}. Our results show that embedding-based selection is the most reliable strategy, avoiding negative transfer and yielding consistent improvements, thereby underscoring the importance of selective augmentation for trustworthy compliance detection.  
\end{itemize}

We made our code implementation and preprocessed data publicly available in this repository~\footnote{https://github.com/farizikhwantri/cross-domain-regcomp}.

\begin{table*}[ht]
\centering
\caption{Compliance Detection as NLI task where GDPR requirement is the premise and Data Processing Agreement is the hypothesis}
\label{fig:example-real}
\begin{tabular}{lp{5cm}p{5cm}ll}
\toprule
Req. & Premise (Req) & Hypothesis (DPA) & DPA-ID & Label \\
\midrule
 R18 & The processor shall assist the controller in consulting the supervisory authorities prior to processing ... to mitigate the risk. (Art. 28(3)(f), Art. 36) & At controller’s request and at the controller’s reasonable expense on a time and materials basis, ... fulfilling any ... Applicable Data Protection Law. & Online 100 &  entailment \\
\bottomrule
\end{tabular}
\end{table*}

\section{Related Work}

\subsection{Compliance detection in Requirements}
Adhering to legal regulations and industry standards has become essential as software systems, particularly in finance and healthcare, become increasingly integrated into everyday life. Software requirements to guarantee compliance must be validated to ensure they align with relevant laws, regulations, standards, and organisational policies, such as those related to data protection (e.g., GDPR, CCPA), cybersecurity (e.g., IEC 62443, NIST 800-53), and safety-critical standards (e.g., HIPAA). This verification process is referred to as compliance detection.

Past studies have thoroughly assessed statistical machine learning and language model techniques for text classification~\cite{azeem2024multi}. Their findings indicate that text classification methods surpass traditional rule-based graph matching. Their method enhanced flexibility and adaptability for managing different types of data. A recent study~\cite{ikhwantri2025explainable} focused on explainable compliance detection by using natural language inference tasks and assurance case structure as a reasoning format.

\subsection{Cross-Domain Data Augmentation}

Data augmentation is a widely used strategy for addressing data scarcity in natural language processing (NLP) tasks. The basic idea is to expand the available training set with additional data, either generated synthetically or transferred from related sources, to improve model robustness and generalisation. In the context of regulatory compliance detection, augmentation is particularly relevant because annotated datasets are both costly to create and highly domain-specific.

Cross-domain data augmentation extends this idea by leveraging data from external but related domains. For example, models trained on GDPR-related texts may benefit from additional data derived from other legal frameworks such as HIPAA or CRA, provided the differences in terminology and reasoning can be managed. However, prior studies in transfer learning have shown that naively adding data from another domain may lead to negative transfer, where the introduced examples degrade rather than improve performance~\cite{pan2010survey, Wang2018CharacterizingAA}. This risk underscores the importance of carefully selecting and filtering cross-domain augmentation data.

Several strategies have been proposed to enhance the effectiveness of cross-domain augmentation. Distribution-based methods, such as Moore–Lewis cross-entropy filtering~\cite{moore-lewis-2010-intelligent}, measure the similarity between candidate examples and the target domain to select relevant samples. Representation-based methods exploit dense embeddings to retrieve semantically aligned data across domains. More recently, large language models (LLMs) have enabled synthetic augmentation~\cite{tan-etal-2024-large}, where new examples are generated to reflect the style and structure of the target domain. Each of these strategies presents distinct trade-offs among diversity, alignment, and scalability.

Within compliance detection, cross-domain augmentation offers the potential to mitigate dataset scarcity and support models that must operate across multiple heterogeneous regulations. However, the effectiveness of augmentation depends critically on how augmentation data is selected and integrated. This problem motivates the need for a systematic evaluation of selection strategies in this setting.

\section{Preliminaries}

This section describes the preliminary method of compliance detection as a natural language inference task.

\subsection{Compliance Detection as Natural Language Inference}

Natural Language Inference (NLI) is a core task in natural language understanding that involves determining the logical relationship between two textual statements~\cite{dagan2005pascal}. We use previous work compliance detection by devising the requirements as premises and evidence as hypotheses~\cite{ikhwantri2025explainable}. This method allows the model to learn from the verbalisation of requirements text instead of discrete classes compared to previous studies, as text classification~\cite{azeem2024multi, ikhwantri2025explainable}. 

For example, in Table~\ref{fig:example-real}, the inputs are GDPR requirements, which served as the premise, and the Data Processing Agreement (DPA) served as the hypothesis. The entailment label means requirements and DPA belong to the same GDPR requirement class and comply, while the non-entailment label means requirements and DPA do not belong to the same GDPR requirement class, i.e. neutral or contradict. 

This approach is similar to requirements mapping for requirements classification approaches~\cite{Fazelnia2024LessonsFT}, which are based on few-shot~\cite{wang2021entailment} or zero-shot scenarios~\cite {sainz-etal-2021-label}. The output is similar to binary classification approaches~\cite{azeem2024multi}. However, NLI only requires a model instead of $n$ models for binary classification, where $n$ is the number of requirements. 

\section{Methodology}

\begin{figure*}[ht]
    \centering
    \includegraphics[width=0.8\linewidth]{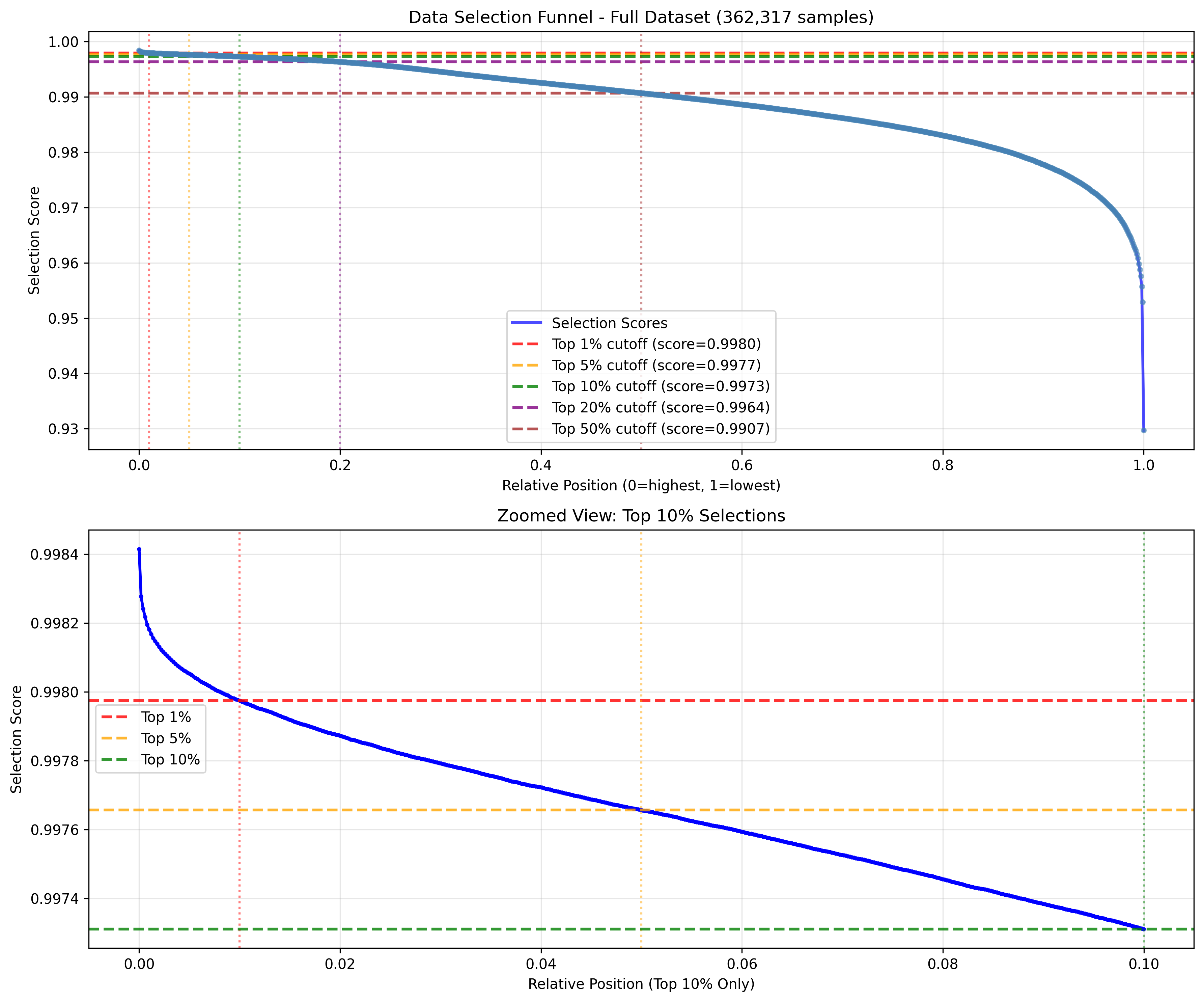}
    \caption{Data selection score and its relative position ranking of embedding similarity method.}
    \label{fig:funnel-embedding}
\end{figure*}


We apply data selection to ensure that the augmented training set contains instances that align closely with the target regulatory domain. Without filtering, irrelevant samples risk introducing noise and causing negative transfer. To address this problem, we evaluate three established strategies from domain adaptation and natural language processing.

\subsection{Moore–Lewis Cross-Entropy Difference (Pre-trained LM-Based Selection)}

The Moore–Lewis method~\cite{moore-lewis-2010-intelligent} used two separate language models, one on the source domain and one on the target domain. For each source instance, the cross-entropy under both models and the difference are computed. Then, the difference between these values is used to rank the source samples as a selection score, with smaller differences indicating stronger alignment with the target distribution.

For simplicity, we use a generic pretrained transformer language model, which is also possible. For each source instance, we compute its cross-entropy under both models and take the difference. We then rank source samples by this difference, where smaller values indicate greater similarity to the target distribution. Finally, we could select the top-ranked subset. 

\subsection{Importance Weighting via Density Ratio}
We estimate how likely a source instance is under the target distribution by computing the density ratio $\frac{p_{\text{target}}(x)}{p_{\text{source}}(x)}$~\cite{xie2023data}. To approximate this ratio, we train a logistic regression classifier that distinguishes target from source samples and transform the predicted probabilities into importance weights. We used the RoBERTa-large encoder to extract text features into the classifier. We then re-weight the source samples during training or filter them by thresholding the ratio. This method provides a statistically grounded way to prioritise source instances that match the target domain.

\subsection{Embedding Similarity}

\begin{figure*}[ht]
    \centering
    \includegraphics[width=0.8\linewidth]{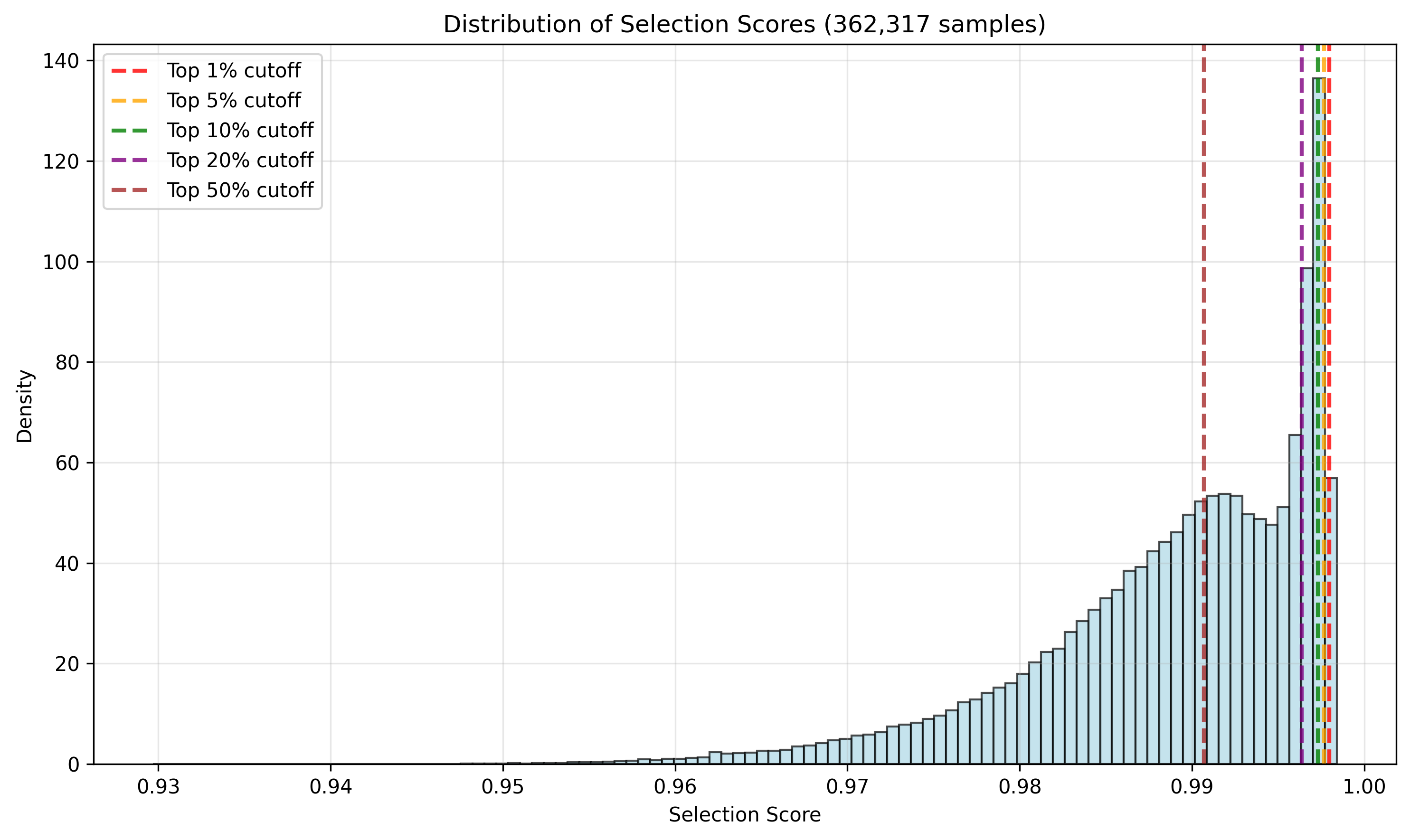}
    \caption{Distribution of embedding similarity of source dataset based on similarity scores to target dataset using RoBERTa-large model}
    \label{fig:dist-embedding}
\end{figure*}

We encode both source and target examples using a pretrained model, such as BERT, and compute their similarity in the embedding space. For each target instance, we retrieve the closest source examples based on cosine similarity. We then select or rank these nearest neighbours as training candidates. This method leverages the representational power of pretrained embeddings and allows us to identify semantically aligned instances directly. Researchers have reported strong results for this strategy in cross-domain NLP and compliance detection tasks. Figures~\ref{fig:funnel-embedding} visualise the embedding similarity score, which is used as a data selection score. The density distribution of the embedding score is shown in Figure~\ref {fig:dist-embedding}.

\section{Dataset}

\subsection{GDPR-DPA}

We use the compliance detection dataset from a previous study~\cite{azeem2024multi} as \textbf{the source domain}\footnote{\url{https://zenodo.org/records/11047441}}. We called this dataset the GDPR-DPA from the General Data Protection Regulation-Data Processing Agreement (GDPR-DPA). The details of the dataset statistics are presented in Table~\ref{tab:cross_domain_dataset_stats}. 
The number of instances in the training set is larger than the original data due to the pairing of negative samples between non-matching premises (requirements) and the hypothesis (DPA). 
We set the sampling rate to 0.1 to add negative pairs within the iteration of unique hypotheses times the number of unique premises.

\begin{table}[ht]
    \centering
    \caption{Dataset statistics for NLI-based compliance detection across source (GDPR-DPA) and target (HIPAA) domains.}
    \label{tab:cross_domain_dataset_stats}
    \begin{tabular}{lrrrr}
        \toprule
        & \multicolumn{2}{c}{\textbf{GDPR-DPA}} & \multicolumn{2}{c}{\textbf{HIPAA}} \\
        \midrule
        \textbf{Statistics} & Train & Test & Train & Test \\
        \midrule
        Instances           & 362,317 & 1,511 & 394 & 1426 \\
        Unique ID           & 61      & 8     & 372 & 10 \\
        Unique premise      & 45      & 45    & 10  & 10 \\
        Unique hypothesis   & 9,820   & 1,364 & 372 & 184 \\
        Unique target       & 45      & 17    & 9   & 1 \\
        \bottomrule
    \end{tabular}
\end{table}

\subsection{HIPAA}

We use the Health Insurance Portability and Accountability Act (HIPAA) dataset in our study as the \textbf{target domain}. The dataset created in 2010~\cite{10.1145/1806799.1806825} and later reused~\cite{guo2017tackling, etezadi2025classificationpromptingcasestudy}. It was manually constructed by identifying trace links between regulatory requirements and regulatory statements derived from the U.S. HIPAA privacy and security provisions.

The dataset covers ten key categories of HIPAA requirements: access control (AC), audit controls (AUD), person or entity authentication (PA), transmission security (TS), unique user identification (UUI), emergency access procedures (EAP), automatic logoff (AL), encryption and decryption (SED), encryption (TED), and integrity controls (IC). Each category reflects critical aspects of data privacy and security within the healthcare domain.

HIPAA consists of ten requirement documents, each composed of shall-statements that express mandatory compliance obligations. In total, the dataset contains 1,891 requirements, of which 243 are explicitly annotated with trace links to regulatory provisions. These annotations provide ground truth for evaluating compliance detection systems by indicating which regulatory statements are satisfied by which requirements.

Table~\ref{tab:cross_domain_dataset_stats} summarises the dataset, presenting the ten HIPAA documents alongside their descriptions and the distribution of trace links across the ten provision categories. This structured format provides both a realistic compliance testbed and an opportunity to evaluate cross-domain transfer.






\section{Experiments Setup}



\subsection{Research Questions} In this paper, we aim to answer the following questions:
\begin{description}

    \item[RQ1] How does cross-domain data selection compare to using only the target domain dataset?
    \item[RQ2] Which data selection method for cross-domain augmentation (random, Moore–Lewis, embedding and importance weighting) leads to the most effective improvements in compliance detection?

\end{description}

\subsection{Settings}

In this study, we use BERT-large~\cite{devlin-etal-2019-bert} (BERT) and RoBERTa-large~\cite{liu2019roberta} (RoBERTa) and Legal-BERT-base~\cite{chalkidis-etal-2020-legal} (Legal-BERT) as the BERT counterpart. We also explore decoder-based LMs, such as GPT-2-XL\footnote{\url{https://huggingface.co/openai-community/gpt2-xl}} and Llama-3 (1B, 3B)~\cite{grattafiori2024Llama} as a text generation baseline for classifying entailment using a prompting method. 

\paragraph{FT-NLI} All models are fine-tuned for six epochs, 64 batch size (32 batch size with two gradient accumulation steps), 1e-05 learning rate, and weight decay 0.01. The models are optimised with a learning rate of 1e-05, and the weights decay by 0.01.

\paragraph{Zero-shot \& One-shot} For Decoder-based LMs, we used zero-shot and one-shot prompts to predict the entailment. The prompt template for zero-shot is:

\begin{centering}
\begin{tcolorbox}[colback=gray!10, colframe=gray, title=Zero-shot Prompt Template, width=0.8\linewidth, fontupper=\footnotesize]
    "Below is a Natural Language Inference (NLI) task for compliance detection in \{regulation\} domain."
    \newline
    "give an answer in either 'entailment' or 'not entailment'"
    \newline
    "Premise: \{premise\}"
    \newline
    "Hypothesis: \{hypothesis\}"
    \newline
    "Answer:"
\end{tcolorbox}
\end{centering}

The prompt template for one-shot is:

\begin{centering}
\begin{tcolorbox}[colback=gray!10, colframe=gray, title=One-shot Prompt Template, width=0.8\linewidth, fontupper=\footnotesize]
    "Below is a Natural Language Inference (NLI) task for compliance detection in the \{regulation\} domain.
    \newline
    \newline"
    "Example 1:\newline
    Premise: \{positive\_example\_premise\}"
    \newline
    "Hypothesis: \{positive\_example\_hypothesis\}"
    \newline
    "Answer: entailment"
    \newline
    \newline
    "Example 2:
    \newline
    Premise: \{negative\_example\_premise\}"
    \newline
    "Hypothesis: \{negative\_example\_hypothesis\}"
    \newline
    "Answer: not entailment"
    \newline
    \newline
    "Premise: \{premise\}"
    \newline
    "Hypothesis: \{hypothesis\}"
    \newline
    "Answer:"
\end{tcolorbox}
\end{centering}

\subsection{Scenarios}

We design our evaluation around two scenarios that reflect practical use cases of compliance detection: in-domain and cross-domain. These scenarios allow us to assess both the effectiveness of data selection strategies within a single regulatory framework and their generalisation across different domains. We report Precision (P), Recall (R), and micro-F1-score (F1).

\paragraph{In-domain}
In the in-domain scenario, we train and evaluate models on the same regulatory framework. For example, we fine-tune an NLI model using the GDPR–DPA dataset and evaluate on held-out GDPR–DPA pairs. This setting captures the common use case where organisations seek to automate compliance verification for a single regulation with sufficient annotated or augmented data. These experiments provide a baseline for evaluating the upper-bound effectiveness of data selection. 

\paragraph{Cross-domain}
In the cross-domain scenario, we transfer models from GDPR–DPA (source) to HIPAA (target), without considering the reverse direction due to data scarcity in the target domain. We evaluate whether models trained on GDPR–DPA clauses generalise to HIPAA trace links. To minimise negative transfer, we employ data selection strategies and vary the proportion of source data utilised. Specifically, we experiment with selection ratios from the set of ${1\%, 5\%, 10\%, 20\%, 50\%, 75\%, 80\%, 90\%}$. This setup allows us to investigate how much carefully chosen source data is necessary to achieve effective transfer, and whether different strategies consistently outperform random sampling at varying levels of data availability. For cross-domain evaluation, we focused on the model which achieved the highest performance on the target dataset.

\section{Results and Discussion}

\subsection{In-domain}

\begin{table}[htbp]
\centering
\scriptsize
\caption{Pre-trained language model trained on Compliance detection 
NLI task.}
\label{tab:finetune-exp}
\begin{tabular}{llcccccc}
\toprule
& & \multicolumn{3}{c}{GDPR-DPA} & \multicolumn{3}{c}{HIPAA}\\
\midrule
Model & Method & P & R & F1 & P & R & F1\\
 \midrule
BERT & FT-NLI & .91 & .81 & .85 & .52 & .55 & .14\\ 

Legal-BERT & FT-NLI & .90 & .83 & .86 & .53 & .57 & .55 \\ 
RoBERTA & FT-NLI & .89 & .72 & .78 & .52 & .61 & .56 \\ 
\midrule
GPT-2-XL (1.5B) & Zero-Shot & .49 & .48 & .48 & .50 & .49 & .46 \\
        & One-Shot & .53 & .58 & .55 & .49 & .43 & .45 \\

Llama-3.2 (1B) & Zero-Shot & .49 & .49 & .49 & .50 & .49 & .42 \\
        & One-Shot & .52 & .53 & .52 & .49 & .47 & .15 \\

Llama-3.2 (3B) & Zero-Shot & .49 & .48 & .48 & .51 & .56 & .20 \\
        & One-Shot & .56 & .52 & .54 & .51 & .57 & .44 \\
 \bottomrule
\end{tabular}
\end{table}

\begin{figure*}[ht]
    \centering
    \begin{tikzpicture}
    \begin{groupplot}[
        group style={
            group size=2 by 1,
            horizontal sep=2cm,  
            vertical sep=1cm,
        },
        grid=both,
        width=0.45\linewidth,
        height=6cm,
        xlabel={\% of selected data},
        ylabel={Macro F1 Score},
        xtick={1, 5, 10, 20, 50, 75, 90},
        xticklabels={1, 5, 10, 20, 50, 75},
        ymin=0.40, ymax=1.0,
        legend to name=sharedlegend_ratio,
        legend columns=5,
        legend style={
            font=\footnotesize,
            draw=none,
            /tikz/every even column/.append style={column sep=1cm}
        }
    ]

    \nextgroupplot

        \addplot[mark=square*, thick, color=blue] coordinates {(1, 0.90) (5, 0.93) (10, 0.87) (20, 0.82) (50, 0.84) (75, 0.90)};
    
        \addplot[mark=triangle*, thick, color=red] coordinates {(1, 0.82) (5, 0.97) (10, 0.75) (20, 0.88) (50, 0.86) (75, 0.92)};
    
        \addplot[mark=diamond*, thick, color=green!] coordinates {(1, 0.91) (5, 0.91) (10, 0.88) (20, 0.85) (50, 0.83) (75, 0.89)};
    
        \addplot[mark=o, thick, color=orange] coordinates {(1, 0.85) (5, 0.89) (10, 0.84) (20, 0.79) (50, 0.81) (75, 0.88)};

        \addplot[thick, dashed, color=black] coordinates {(1, 0.84) (10,  0.84) (20,  0.84) (50,  0.84) (75,  0.84) (100, 0.84)};

    \nextgroupplot
        \addlegendimage{mark=square*, color=blue}
        \addlegendentry{Embedding}
        \addlegendimage{mark=triangle*, color=red}
        \addlegendentry{Importance Weighting}
        \addlegendimage{mark=o, color=orange}
        \addlegendentry{Random}
        \addlegendimage{mark=diamond*, color=green!}
        \addlegendentry{Moore-Lewis}
        \addlegendimage{dashed, color=black}
        \addlegendentry{Full}

        \addplot[mark=square*, thick, color=blue] coordinates {(1, 0.61) (5, 0.65) (10, 0.64) (20, 0.57) (50, 0.86) (75, 0.62)};
    
        \addplot[mark=triangle*, thick, color=red] coordinates {(1, 0.71) (5, 0.81) (10, 0.67) (20, 0.75) (50, 0.85) (75, 0.63)};
    
        \addplot[mark=diamond*, thick, color=green!] coordinates {(1, 0.76) (5, 0.73) (10, 0.81) (20, 0.86) (50, 0.87) (75, 0.71)};
    
        \addplot[mark=o, thick, color=orange] coordinates {(1, 0.68) (5, 0.69) (10, 0.78) (20, 0.58) (50, 0.71) (75, 0.62)};

        \addplot[thick, dashed, color=black] coordinates {(1, 0.70) (10,  0.70) (20,  0.70) (50,  0.70) (75,  0.70) (100, 0.70)};

    \end{groupplot}
    \end{tikzpicture}

    \ref{sharedlegend_ratio}

    \caption{F1 Scores of RoBERTa Models of increasing selected ratio from 1\% to 75\% across 3 data selection methods and a random baseline. Left: \textbf{Model performance on validation set (held from training model)} but which overlap with data selection. Right: \textbf{Model performance on test data}.}
    \label{fig:data_selection-result}
\end{figure*}
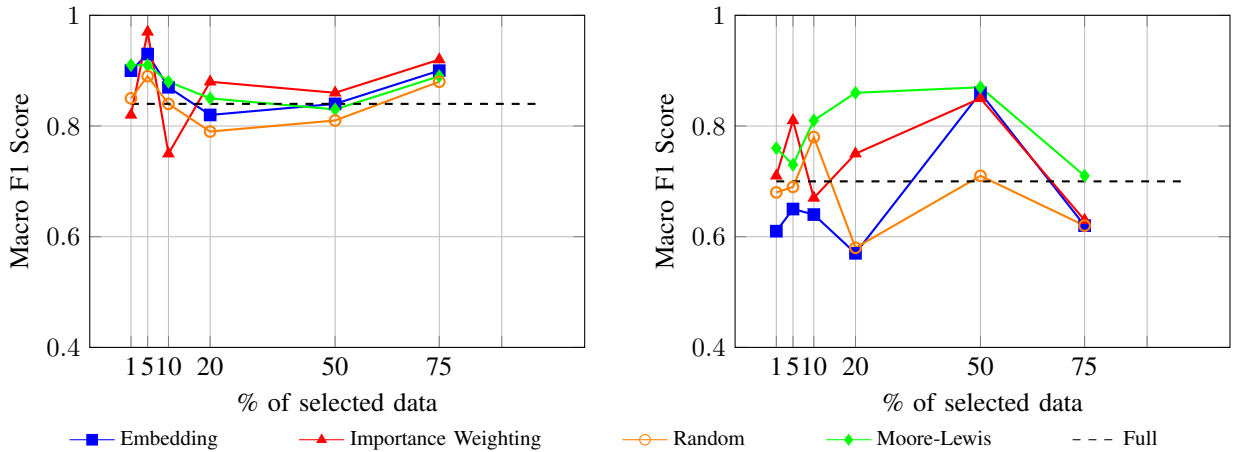

Table~\ref{tab:finetune-exp} presents the in-domain performance of pre-trained language models on the compliance detection task, evaluated on both the GDPR-DPA and HIPAA datasets. Among the fine-tuned NLI models, Legal-BERT achieves the strongest overall balance on GDPR-DPA with an F1 of 0.86, followed by BERT at 0.85 and RoBERTa at 0.78. These results suggest that domain-specific pretraining (Legal-BERT) offers a significant benefit when applied to a legal-domain dataset, such as GDPR-DPA. 

In contrast, performance on HIPAA is consistently lower across all models, with F1 scores ranging between 0.54 and 0.59. RoBERTa delivers the best recall on HIPAA (0.61), whereas Legal-BERT provides the highest overall F1 score (0.56). These results indicate that models trained on GDPR-DPA generalise poorly to HIPAA, reflecting the challenge of adapting to domain-specific phrasing, terminology, and regulatory intent. 

Decoder LLMs evaluated in zero-shot and one-shot setups consistently underperform compared to fine-tuned encoders. GPT-2-XL, Llama-3.2 (1B), and Llama-3.2 (3B) hover near random baselines in zero-shot settings (F2 \~0.48–0.49). One-shot prompting slightly improves results, with Llama-3.2 (3B) achieving an F2 of 0.54 on GDPR-DPA and 0.56 on HIPAA. These improvements, however, remain well below the fine-tuned discriminative baselines. These results highlight the current limitations of generative LLMs for highly specialised compliance tasks when used with minimal supervision.

Taken together, these findings emphasise two key points. First, supervised fine-tuning on compliance-specific data yields significantly stronger results than zero- or few-shot prompting with large generative models. Second, even with fine-tuned models, a significant performance gap remains between the source (GDPR-DPA) and the target HIPAA. This gap motivates the exploration of targeted data selection and augmentation methods to improve cross-domain robustness in compliance detection.

\subsection{Cross-domain}

\begin{figure*}[ht]
    \centering
    \includegraphics[width=0.8\linewidth]{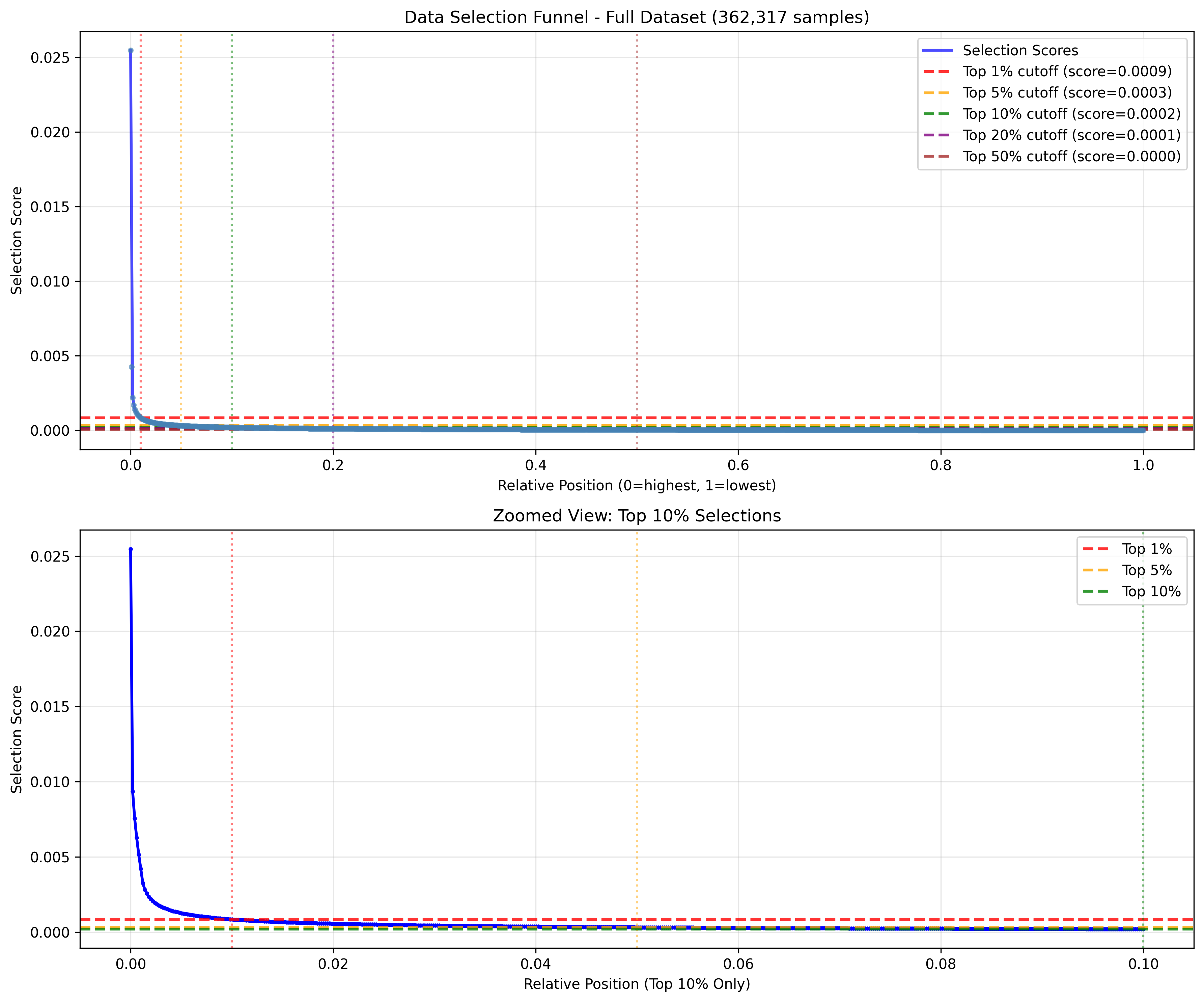}
    \caption{Data selection score and its relative position ranking of importance weighting method.}
    \label{fig:funnel-importance}
\end{figure*}

We evaluated the impact of increasing the proportion of selected source-domain data on held-out validation and test performance using RoBERTa. Figure~\ref{fig:data_selection-result} summarises results across three selection strategies: embedding similarity, importance weighting, and Moore–Lewis cross-entropy difference against two baselines, random sampling and full augmentation.

On the validation set, all data selection strategies showed clear performance variations as the ratio of selected data increased. Embedding-based selection achieved strong results at both very low (1\%) and high (75\%) ratios, with F1 scores of 0.90, demonstrating robustness across extremes. Importance weighting delivered the highest peak performance (0.97 at 5\%), but also exhibited instability, with a substantial drop at 10\%. Moore–Lewis filtering remained relatively stable, achieving competitive performance across all ratios, though it lacked the sharp peak observed in importance weighting. The random baseline performance lagged but followed a similar trend of performance decline at intermediate ratios compared to targeted methods. The baseline upper-bound full augmentation is 84\%. These baseline results suggest the presence of negative transfer when larger amounts of less-relevant source data were included.

We observe that using only the top 1\% of selected data already yields strong performance across methods. For example, embedding similarity achieves an F1 of 0.90, Moore–Lewis reaches 0.91, and importance weighting obtains 0.82. These results highlight that carefully chosen small subsets of data can rival or surpass performance obtained with much larger proportions, underscoring the efficiency of selective augmentation.

When comparing performance across selection ratios, the highest overall accuracy is achieved by importance weighting at 0.97 with 5\% data, outperforming other methods at similar scales. This performance suggests that density-ratio-based selection is particularly effective in aligning source and target distributions when sufficient overlap exists between them. Embedding similarity also performs robustly, peaking at 0.93 with 5\% data, while Moore–Lewis remains stable around 0.91. In contrast, random sampling produces weaker results at the same scale, peaking at 0.89.

We detect signs of negative transfer at intermediate ratios. For example, embedding similarity drops from 0.93 at 5\% to 0.82 at 20\%, and importance weighting falls sharply from 0.97 at 5\% to 0.75 at 10\%. These degradations indicate that adding poorly matched source samples introduces noise, harming performance. However, performance tends to recover at higher ratios (e.g., 0.90 for embedding similarity at 75\%, 0.92 for importance weighting at 75\%), likely due to the model leveraging a broader and more representative pool of data once enough “pure” signal dominates.

Overall, these findings suggest that small, carefully selected subsets of source data are often sufficient and superior to larger, less filtered pools. Among the methods, importance weighting shows the strongest peak performance, while embedding similarity provides stable gains with a lower risk of overfitting to mismatched samples. These results underscore the importance of principled data selection in mitigating negative transfer for cross-domain compliance detection.

On the test set, the differences between methods became more pronounced. Importance weighting again reached the strongest peak at lower ratios (0.81 at 5\%), but its performance sharply declined at larger selections (0.63 at 75\%). Moore–Lewis filtering demonstrated the most consistent gains, climbing steadily from 0.73 to 0.87 between 5\% and 50\%, and outperforming other methods at higher selection levels. Embedding-based retrieval, while strong on validation, suffered from volatility on test performance, with particularly poor generalisation at 20\% and 75\%. Random selection remained weaker overall, with no ratio providing clear advantages over targeted methods, and the full augmentation baseline is 70\% F1 score.

\begin{tcolorbox}[colback=white, colframe=black, boxrule=0.5pt, arc=4pt, left=4pt, right=4pt, top=2pt, bottom=2pt]
\textbf{RQ1: The cross-domain data selection improves the model performance compared to using only the target domain. } 
\end{tcolorbox}

These results highlight two key observations. First, negative transfer is evident when ratios of 10–20\% are used, where most methods reach their lowest F1 scores before recovering at higher levels, indicating the disruptive effect of partially aligned data. Second, the effectiveness of a selection method depends on both the data ratio and evaluation perspective: importance weighting excels at low-ratio scenarios but struggles with scaling. At the same time, Moore–Lewis provides the most reliable improvements at larger ratios. Overall, these findings suggest that mitigating negative transfer in cross-domain compliance detection requires not only careful method choice but also tuning of the proportion of source data introduced.

\begin{tcolorbox}[colback=white, colframe=black, boxrule=0.5pt, arc=4pt, left=4pt, right=4pt, top=2pt, bottom=2pt]
\textbf{RQ2: Importance weighting and Moore-lewis emerged as the most effective, outperforming the random baseline.}  
\end{tcolorbox}

\begin{table*}[ht]
\centering
\caption{Top-ranked example (0-10\%) shows meaningful domain security concept transfer. Later ranked (10-20\%) example illustrates overfitting to generic legal template phrasing, high lexical scores despite semantic divergence.}
\label{tab:qualitative-example}
\begin{tabular}{lllp{4.0cm}p{4.0cm}ccc}
\toprule
Interval & Channel & Source Rank & Source Text & Target Text & Jaccard & BLEU-2 & ROUGE-L \\
\midrule
0–10\%  & hyp→hyp & 8489 & “Inactive session are shut down   after a defined period of inactivity.” & “Automatic logoff. Implement   electronic procedures that terminate an electronic session after a   predetermined time of inactivity.” & .2381 & .0711 & .3704 \\
10–20\%  & hyp→prem & 39744 & “An Incident may include but not   be limited to:” & “Patient information may include   (but is not limited to) historical patient documentation and test results.” & .3529 & .1199 & .5000 \\
\bottomrule
\end{tabular}
\end{table*}

\subsection{Negative Transfer Analysis}

Early, high-ranked source slices (top 5\%--10\%) yield moderate lexical overlap (e.g., hyp$\rightarrow$hyp Jaccard $\approx 0.24$, ROUGE-L $0.37$ for session timeout clauses) that corresponds to genuine semantic alignment with target HIPAA security controls (automatic logoff, user authentication, transmission security). These pairs offer actionable cross-domain signals by matching operational intent (access control, integrity, encryption) rather than only surface wording. We measure set-level overlap statistics, such as source hypothesis to target premise with $0.21$ Jaccard, with substantial target token coverage. This metric shows that a small, curated fraction already captures most domain-relevant vocabulary. Table~\ref{tab:qualitative-example} shows one of the examples of positive transfer on top-ranked (0-10\%) and negative transfer on 10-20\% source data to the target text premise. 

In contrast, extending selection introduces clauses whose elevated lexical scores are driven by generic regulatory templates (``may include but is not limited to'', enumerative appendix headings, boilerplate confidentiality statements). Later hypothesis-premise examples show higher ROUGE-L (up to $0.53$) and Jaccard ($>0.35$) without proportional semantic specificity, indicating a lexical–semantic decoupling. This drift risks reinforcing superficial patterns (enumeration frames, modal obligation scaffolds) rather than precise compliance predicates.

Overall, the results support capping augmentation at an early percentile and prioritising semantically grounded retrieval (importance weighting or Moore-Lewis entropy) over bulk inclusion. Monitoring divergence between lexical overlap (Jaccard/ROUGE) and functional alignment (domain control match) provides an operational criterion for halting further data addition to avoid negative transfer.

\section{Threats to Validity}


Our study investigates cross-domain data augmentation and selection for compliance detection. While the experiments demonstrate promising improvements, several validity concerns remain.

\textbf{Construct validity}. Our evaluation assumes that selection metrics (e.g., embedding similarity, density ratios, cross-entropy difference) faithfully reflect the relevance of source-domain data to the target domain. However, these metrics may not capture regulatory nuance, such as legal context or domain-specific terminology shifts between GDPR and HIPAA. We addressed this by testing selection strategies across a range of ratios, identifying consistent patterns such as negative transfer at intermediate scales. Future work should incorporate compliance-aware representations or domain-specific pretrained LMs to strengthen construct validity.

\textbf{Internal validity}. A key risk lies in whether improvements arise from genuine cross-domain adaptation or from dataset artefacts. Selection methods such as embedding similarity or Moore–Lewis may inadvertently prioritise superficial lexical overlap rather than deeper semantic alignment. To mitigate this, we compared multiple data selection baselines (random, embedding, importance weighting, and Moore–Lewis) under identical settings. A further mitigation would involve ablation studies to separate performance gains due to true semantic transfer from those due to surface-level bias.

\textbf{External validity}. Our experiments transfer from GDPR-DPA to HIPAA, providing evidence for cross-domain performance under a significant distributional shift. However, results may not generalise to other compliance domains such as finance, safety-critical regulations, or cybersecurity standards. Moreover, we tested transfer in a single direction (GDPR → HIPAA). The main issue is that the target dataset differs substantially in terms of quantity. Nonetheless, further experiments on additional domains and bidirectional transfers are needed to confirm broader applicability.


\section{Conclusion}

This study examined the role of cross-domain data selection in compliance detection through two research questions.  

\textbf{RQ1} asked how cross-domain data selection compares to using only target-domain data. The results indicate that incorporating carefully selected subsets of source-domain data consistently improves performance relative to target-only training. This finding confirms that transfer across domains is beneficial, provided that mismatched samples are effectively controlled.  

\textbf{RQ2} addressed which selection method yields the strongest improvements. Importance weighting and Moore-Lewis emerged as the most effective, outperforming the random baseline. These methods reduced negative transfer by filtering out poorly aligned data, showing that targeted selection contributes more than simply increasing data volume. However, performance degradations at intermediate selection ratios also revealed the sensitivity of compliance detection models to noise introduced by less compatible samples.  

Future work should extend beyond selection toward controlled augmentation, including methods such as paraphrasing~\cite{wu2019conditional, li2022data, sharma-etal-2023-paraphrase}, backtranslation~\cite{sugiyama-yoshinaga-2019-data}, and structured generation through assurance case–guided formats~\cite{ikhwantri2025explainable}. Further analysis of cross-domain transfer of models’ internal representations~\cite{du2021learning, 10.1145/3623399} and their explanation faithfulness~\cite{steen-etal-2023-little, 10.1145/3653984} will also be essential to improve trustworthiness and transparency. Improving along these directions may enable compliance detection systems that generalise more robustly across regulatory domains while providing reliable justifications for their predictions.

\section*{Acknowledgements}

This work has been funded by the European Commission under grant agreement No. 101120606, the CERTIFAI project. This work has also benefited from the Experimental Infrastructure for Exploration of Exascale Computing (eX3), which is financially supported by the Research Council of Norway under contract 270053.

\bibliographystyle{IEEEtran}

\bibliography{ieee_custom}

\end{document}